\documentclass[letterpaper]{article}
\usepackage{aaai}
\usepackage{hyperref}
\usepackage{epsfig}
\usepackage{amssymb}
\usepackage{amsmath}
\usepackage{amsfonts}
\usepackage{xspace}
\usepackage{tikz}
\usetikzlibrary[graphs]
\usepackage{algorithm}
\usepackage[noend]{algorithmic}
\usepackage{times}
\usepackage{helvet}
\usepackage{color}
\usepackage{courier}
\usepackage{graphicx}
\usepackage{framed}
\newcommand{\ignore}[1]{}
\usepackage{listings}
\usepackage{todonotes}

\lstdefinelanguage{scala}{
  morekeywords={abstract,case,catch,class,def,%
    do,else,extends,false,final,finally,%
    for,if,implicit,import,match,mixin,%
    new,null,object,override,package,%
    private,protected,requires,return,sealed,%
    super,this,throw,trait,true,try,%
    type,val,var,while,with,yield
    },
  otherkeywords={=>,<-,<\%,<:,>:,\#,@},
  sensitive=true,
  morecomment=[l]{//},
  morecomment=[n]{/*}{*/},
  morestring=[b]",
  morestring=[b]"""
}
\newcommand{\x}{\boldsymbol{x}}
\newcommand{\y}{\boldsymbol{y}}

\newcommand{\Saul}{\textit{Saul}\xspace}
\newcommand{\datamodel}{\textit{data-model}\xspace}
\newcommand{\datareader}{\textit{data reader}\xspace}
\definecolor{orange}{RGB}{255,112,0}
\definecolor{dkgreen}{RGB}{70,148,89}

\begin{document}

\title{
Relational Learning and Feature Extraction by \\ Querying over Heterogeneous Information Networks}

\author{Parisa Kordjamshidi~\footnotemark[2]~ {\bf Sameer Singh}\footnotemark[4]~ Daniel Khashabi\footnotemark[3] 
        ~ Christos Christodoulopoulos\footnotemark[5]
        \\{\Large
        {\bf Mark Summons}\footnotemark[3]
      ~ {\bf Saurabh Sinha}\footnotemark[3]
      ~ {\bf Dan Roth}\footnotemark[3]} \\
        \footnotemark[2]~Tulane University ~ \footnotemark[3]~University of Illinois at Urbana-Champaign ~
        \footnotemark[4]~University of California, Irvine \\
        \footnotemark[5]~Amazon Research Cambridge, UK \\
        \footnotemark[2]~{\tt pkordjam@tulane.edu} ~ \footnotemark[4]~{\tt sameer@uci.edu} ~ \footnotemark[5]~{\tt chrchrs@amazon.co.uk} \\
        \footnotemark[3]~{\tt\{khashab2,mssammon,sinhas,danr\}@illinois.edu }
}

\maketitle

\begin{abstract}

Many real world systems need to operate on heterogeneous information networks that consist of numerous interacting components of different types. Examples include systems that perform data analysis on biological information networks; social networks; and information extraction systems processing unstructured data to convert raw text to knowledge graphs.  Many previous works describe specialized approaches to  perform specific types of analysis, mining and learning  on such networks. In this work we propose a unified framework consisting of a data model -a graph with a first order schema- along with a declarative language for constructing, querying and manipulating such networks in ways that facilitate relational and structured machine learning. 
In particular, we provide an initial prototype for a relational and graph traversal query language where queries are directly used as relational features for structured machine learning models.  Feature extraction is performed by making declarative graph traversal queries.  Learning and inference models can directly operate on this relational representation and augment it with new data and knowledge that, in turn, is integrated seamlessly into the relational structure to support new predictions. We demonstrate this system's capabilities by showcasing tasks in natural language processing and computational biology domains.
\end{abstract}

\section{Introduction} 

Many real world systems need to operate on heterogeneous information networks~\cite{DBLP:journals/corr/ShiLZSY15} that consist of multiple interacting components of various types. Examples include biological networks containing genes and proteins along with experimental genomic and clinical data of the patients; social networks, such as citation networks relating authors and papers; or even more complex networks such as knowledge graphs, which can contain a large variety of types of entities and relationships~\cite{Nickel2015ARO}.   


Although previous research extensively addresses the challenges of working with such networks for various mining tasks (see for example~\cite{Sun:2013:MHI:2481244.2481248,DBLP:conf/edbt/KuckZYCH15}), a general solution for easily constructing such networks or systematically manipulating them by various analysis units has not yet been worked out. In other words, mostly specialized approaches are proposed to perform specific types of analysis over a network design and the implementations are mostly dependent on the type of tasks and the type of analysis (for an overview, see~\cite{DBLP:journals/corr/ShiLZSY15}).


Data representation and flexible intelligent data analysis, as well as the evolution of these networks based on the analysis outcomes, need to be placed in a well-defined framework. A complex version of such networks are instantiations of \emph{knowledge graphs} and as Nickel, et al. put it: ``Representing, learning, and reasoning with [knowledge graphs] remains the next frontier for AI and machine learning.''~\cite{Nickel2015ARO}.
In this work, we move closer to this frontier by: 1) proposing a unified framework for integrating data from heterogeneous resources in one relational graph structure; 2) proposing a query language for constructing, manipulating and evolving this graph; 3) providing the capability of performing relational machine learning and feature extraction on this graph using \emph{the same query language}. The novelty of our work, \emph{which is in progress}, is in the integration of the learning-based analysis with the above components in one system, and allowing the proposed graph query language to be used consistently for preparing learning examples, extracting relational features, and processing the results of the learning models.

Our proposal is different from the conventional usage of query languages in the context of information networks, which are either standard retrieval queries~\cite{He:2008:GQL:1376616.1376660} or designed for a specialized task~\cite{DBLP:conf/edbt/KuckZYCH15}. 
Here, relational and structured machine learning models are declared using a succinct definition language, and are directly applied on the graph. The resulting predictions can be integrated into the same graph in a seamless manner. 
Our model can be seen as an information extraction model that can make declarative queries from unstructured data by expressing them in a relational graph structure. This combines the aspirations of existing works that have tried to combine information extraction modules with relational database systems and use standard querying languages for retrieving the  information~\cite{Krishnamurthy:2009:SSD:1519103.1519105} with those of systems that are designed for processing textual data and which provide a regular expression language to directly query from text~\cite{fextor}. Comparatively, our first order graph query language provides the flexibility and expressivity to extract relational and global patterns from unstructured data for various kinds of data analysis, including feature extraction as well as basic search and retrieval from the network. 

This query language is designed as a part of \Saul,\footnote{\Saul code: \url{https://github.com/CogComp/saul}} a declarative learning based programming language~\cite{KordjamshidiRoWu15}. The queries are handled by Scala's underlying collection manipulation and query optimization is out of the scope of this paper though it is an important issue for future consideration.
Using \Saul for structured learning in NLP domain has been investigated in~\cite{KKCMSR16}. Here, focusing on the data modeling aspect we show that our model facilitates working with data coming from heterogeneous resources using NLP as well as computational biology applications. 
We provide a detailed application of our system for construction and manipulation of biological networks~\cite{BioNetwork}, where the biologist needs to assess patient gene influence on patient drug response, using genomic data. 
Using our language, the biologist can specify their problem in a few lines of code: making declarative queries to perform various analysis. 


\begin{figure}
\centering
\includegraphics[trim=0.4cm 0.5cm 0cm 1.3cm, scale=0.13]{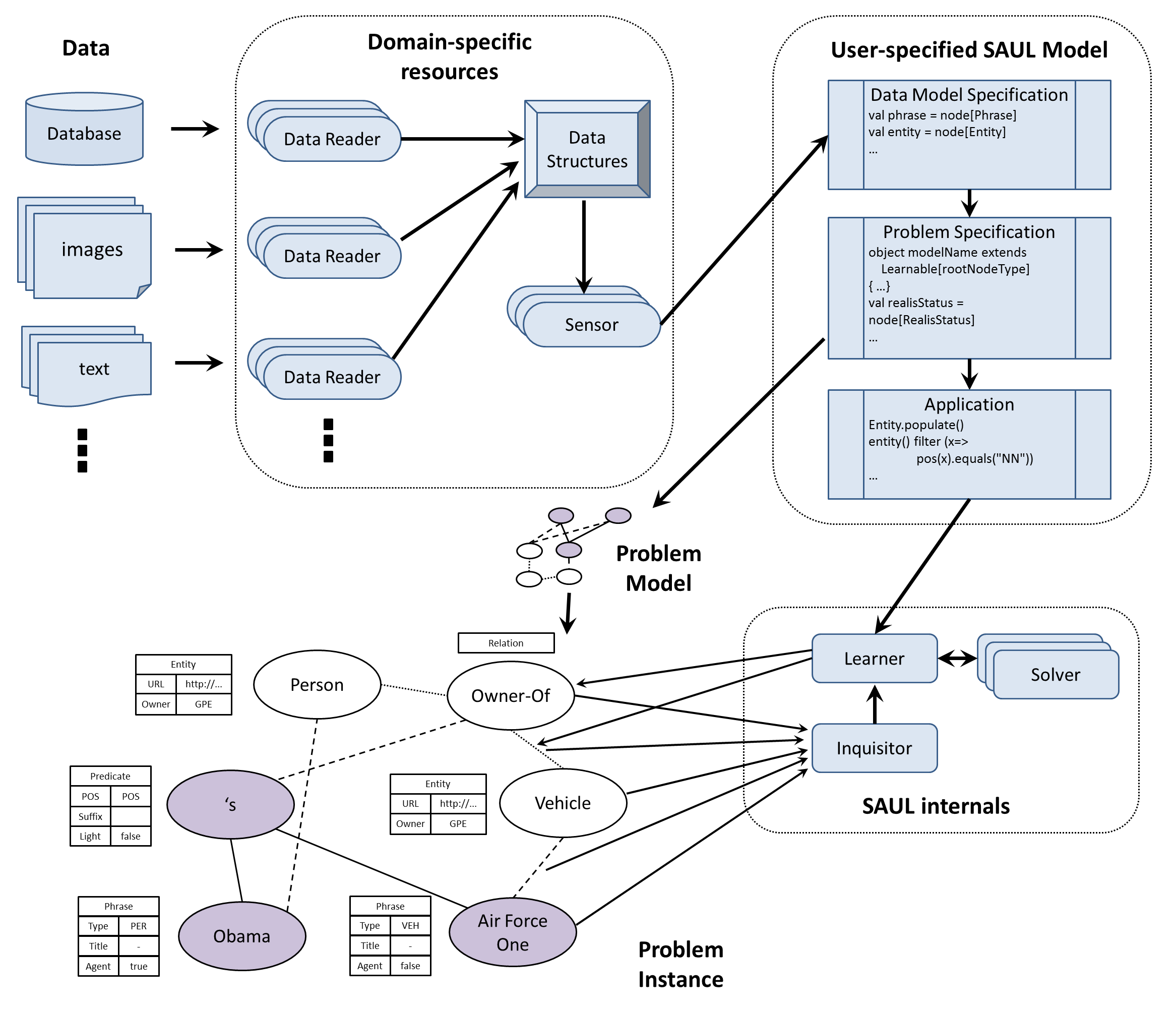}
\caption{Components' interaction with the language.}
\label{fig:overview}
\end{figure}

\section{System overview}
\label{sec:overview}
Figure~\ref{fig:overview} shows the components of the system and their interaction with declarations in the language. \textbf{Domain-specific resources} refer to data structures and available programs that read data files into those data-structures along with any available external functions that can process the data. \textbf{\Saul\ internals} are system components for learning algorithms, inference solvers and internal graph representation for computing the queries which are not exposed to the user. The \datamodel\ specification is the user's schematic specification of the conceptual model of the data domain which is then used to specify the inputs and outputs relevant to a given problem. This is used as a template to generate problem instances. \textbf{Problem instances} are graph data-structures populated with the actual input data. The user-specified \Saul\ model consists of three specification blocks of code: one for the data model, one for the problem specification and domain knowledge, and one for the application that loads the data and applies the learning algorithms/solvers.

In a complete system, the data is read using programs called \emph{\datareader{s}}. The user declares the schema of the data as a graph in which data items are collections of nodes, edges and properties, which are defined using \emph{Sensors}, succinct specifications in \Saul's definition language. The data from the reader is mapped into the \datamodel and a populated graph containing all the data is generated. After population, all the operations such as example generation, feature selection, querying, and learning are defined using user-specified graph queries expressed in terms of \datamodel components. The predictions of the learning models can be incorporated into the \datamodel and the outcomes are integrated into the generated data graph, allowing them to be queried by the feature extraction and learning components. Basically, these components are used to put the raw data into the graph structure and capture domain-specific knowledge, including learned representations of the data itself (e.g. a collection of raw image data can be used to train to recognize an object node).

\section{Data Modeling}

\label{ssec:reading}

 One main goal of our framework is to facilitate using heterogeneous information from various domains in a unified framework.  For example, raw data can be textual documents, a collection of images or videos or spreadsheets describing patients, drug responses. The basic notions that we use including  \emph{readers, baseTypes, sensors, etc} and using these to build an arbitrary graph structure provide an abstraction that paves the way for achieving this goal.  

The \datamodel is a graph schema that is used to explicitly represent the structure of the data and contains \emph{typed} nodes, edges and properties. The node types are domain's basic data-structures, called \emph{base-types}. The base-types are pre-established for each ap plication domain. The \Saul base-types for NLP domain are discussed in~\cite{KKCMSR16} and include for example base-types for representing documents, sentences, phrases, etc, referred to as \emph{linguistic units}. Our computational biology \datamodel is augmented by necessary base-types to represent genes, patients, etc. \footnote{Data from the KnowEng: \url{http://www.knoweng.org/}} 

We use the notion of \datareader{s} which are programs that can read the data into the base-types. Each domain is equipped with a set of \emph{sensors} that are black-box functions applied on the base-types and can generate new nodes, connect them by edges or generate properties of the nodes. 
The graph schema is a first order graph that is based on types of nodes instead of representing all individual objects. The programmer declares the schema of the data, which later, will be populated with the actual data instances in the program and used for feature extraction and learning. 
A node of type \lstinline{T} is declared as in line 1 below and a property for node {\tt nodeName} is declared as in line 2:
\begin{lstlisting} 
val nodeName = node[T]
val propName = property(nodeName) {/*body*/}
\end{lstlisting}
In the \lstinline{body} of the property declaration we use a \emph{sensor} that operates on type \lstinline{T} and returns a value of \emph{property type}. 
The property type can take any of Scala's basic data types 
or a standard collection. 
\noindent A \emph{property sensor} associates each object with a property type. \noindent An edge that connects a source node \lstinline{node1} of type \lstinline{T} to a destination node \lstinline{node2} of type \lstinline{U} is defined as {\tt edge(node1,node2)}.  
\noindent The edge declaration defines the type of edge; sensors will populate the instances with actual edges. \noindent A sensor that defines how a node is connected to another node is called an \emph{edge sensor}. 
We can easily add sensors to edges as \lstinline{edgeName.addSensor(sensorName)}. 
\noindent Edge sensors can be of two types, \emph{matching} or \emph{generating}. 

\noindent{\bf Matching sensors} are Boolean functions that establish an edge between two existing nodes if certain conditions hold. 
The matching edge sensors can be easily defined based on specific properties that source and destination nodes possess.

\noindent{\bf Generating sensors}, given a source node, create a number of destination nodes and at the same time establish a connection between the source and destination. For example, a tokenization edge, generates tokenizated token nodes, upon the population  of the sentence nodes. 
The edges establish connections in both directions automatically and provide the flexibility of working with a relational model. 
 

\begin{algorithm}
\caption{\label{alg1} { \small 
Declaring schema-graph $M$ with base-types $B$ } }
\begin{algorithmic}[1]
\FORALL {$b \in B$}
    \STATE Define a node in $M$ as 
     \lstinline{val n}$_i$ \lstinline{= node[}$b$\lstinline{]}
    \STATE Define properties of $b$ based on the available property sensors $sense_b$ as \\ 
     \lstinline{val  p}$_i$ \lstinline{ = property(n}$_i$\lstinline{)} { \tt   \{ sense}$_b${\tt  \}}
\ENDFOR
\FORALL {$n_i$ and $n_j$ $\in B$ which you need to establish a connection}
\STATE {Define an edge and add edge sensors to it: } \\ 
\lstinline{val e}$_i$ {\tt = edge(}$n_i,n_j${\tt ) \{ sense}$_{r(b_i,b_j)}${  \tt  \}}
\ENDFOR
\end{algorithmic}
\end{algorithm}

\noindent Algorithm~\ref{alg1} shows the steps needed for declaring the \datamodel. For example, a typical NLP \datamodel can have the following nodes and edges,
\begin{lstlisting}
val sentences=node[Sentence]
val phrases=node[Phrase]
val relations=node[Relation]
val phraseToRelations=edge(phrase,relations)
...
\end{lstlisting}
\subsection{Populating the data model}
Once a \datamodel\ is specified, we can populate it with the actual data instances to get a \emph{propositionalized} data graph. 
For example, given a \datareader that provide collections of objects for all declared nodes, we can populate the node, \lstinline{nodeName} as: \lstinline{nodeName.populate(collectionName)}
The edges between nodes are made automatically using matching sensors. For example \lstinline{sentences.populate(sentenceList)} will populate the sentence nodes with the list of sentences provided by the \datareader.
When populating a node if the necessary generating sensors are added to edges, then populating sentences can generates tokes, phrases, etc. 

\section{Graph Queries}\label{Graph Queries}
The \Saul\ \datamodel helps to explore, query, and use the data to design relational features for various learning models. Having a first order graph has the advantage of enabling queries from both the schema of the graph (for searching meta patterns like meta-path~\cite{Sun:2012:MHI:2371211} features), as well as from instances of the populated data graph.  

The queries take advantage of both the object-oriented and the functional programming paradigms in Scala.  
The queries can be applied on individual nodes or on collections of them and can return properties, nodes, or collections of those. 
Accordingly, the return types of the queries will be basic Scala types, base-types, or collections of those.   
A major advantage of our approach is that we can \emph{chain} different commands together to form complex queries, which are supported by Scala's powerful and expressive syntax and its ability to handle such compositions.

A basic query can be composed of a node name as \lstinline{NodeName()}, which returns a collection of all data instances of that node; calling an edge as \lstinline{$\sim$>EdgeName}\footnote{$\sim >$ is \Saul's traversal function} to \emph{follow} the called edge; and calling a property by \lstinline{prop PropertyName} when it is applied on a collection, or by \lstinline{PropertyName(x)} when \lstinline{x} is a single object. 

\subsection{Relational algebraic operations}

By analogy to databases, each node can be seen as a relational table that describes a collection of objects or the relationships between objects. 
Given this perspective, performing relational algebra operations is equivalent to manipulating the collections in the nodes. 


\noindent{\bf Graph traversal}. The relational join operations that collect information across nodes are performed with graph traversal queries.
To clarify this, assume that we need to get all relations connected to  a phrase x, we start from an instance \lstinline{x} of \lstinline{phrases} node and follow an edge type named \lstinline{phraseToRelations} to get all relations connected to \lstinline{x}, that is, \lstinline{phrases(x)$\sim$>phraseToRelations}. In general, we use \lstinline{$\sim$>edgeName} to retrieve objects connected to either a single object \lstinline{x}, or a collection \lstinline{c()}. The \lstinline{c()$\sim$>edgeName} is equivalent to applying \lstinline{edgeName} to every single element in \lstinline{c}. 
\Saul's data model establishes a reverse edge automatically given each edge declaration, e.g.,
the reverse access is made by the expression \lstinline{$\sim$>-phraseToRelation}.  
 We can extend the above query to get all phrases that are connected to the source phrase \lstinline{x} by means of some \lstinline{relations} node as follows:
\begin{lstlisting}
val connectedPhrases = phrase(x)$\sim$> phraseToRelation$\sim$>-phraseToRelation
\end{lstlisting}
{\bf Explicit joins.} Though the graph traversal operations provide the functionality that is expected from classic relational join operations, the user can join nodes explicitly and add complex relational nodes to the graph. This can be done as follows: 
\begin{lstlisting}
val joinNode = join(node1,node2)(/*body*/)
\end{lstlisting}
where the body is a logical expression based on the properties of the nodes that indicates which node instances should be joined together. For example, \lstinline{(_.posTag == _.posTag)}\footnote{In Scala, the \lstinline{_} acts as a placeholder for parameters in the anonymous function.} indicates that we need to join the nodes which have the same posTag property values. The outcome node is represented as a tuple of the two nodes and gives access to all their properties, edges and instances.  

\noindent{\bf Filtering.} The Scala's \lstinline{filter} is used to \emph{Select} a set of nodes that meet a specified condition. For example, \lstinline{words().filter(x=>pos-tag(x).equals("NN"))} selects a set of \emph{words} nodes whose \lstinline{pos-tag} is NN.
\subsection{Pattern-matching and graph-isomorphisms}

The \datamodel provides the advantage of finding patterns of data instances as well as meta patterns according to the type of nodes and edges in the data graph or its schema i.e. \datamodel. These kinds of queries are not easy to formulate in standard relational data models. 

\noindent{\bf Contextual queries}. We provide functions that help explore the context of a node and its neighborhoods for more flexible pattern matching and accessing global patterns. The edges provide only one step access from one node to another node that is directly connected to it. The command \lstinline{node(x).neighborAt(n)} gives the collection of nodes that are exactly \lstinline{n} edges away from node instance \lstinline{x}.
We also provide a \lstinline{neighborWithin(n)} variant that provides the collections of nodes that are at least \lstinline{n} edges away, and for both these functions the users can include an optional set of edges to restrict the neighborhood by.  
Like the rest of the queries, neighborhood queries allow composition via chaining: it can be applied to a result of another query, and other operations such as filtering, aggregation, or traversal can be applied on its result. \\
\noindent{\bf Path queries}. 
In order to identify the shortest path between two data instances, we provide the \lstinline{path} function that can be applied to node queries, \lstinline{node(x).path(y)}. In particular, this function searches for an instance \lstinline{y} that is reachable from a path of edges starting from instance \lstinline{x}, and returns the sequence of edges that connect these two nodes (which is empty if a path does not exist). 
Optionally, the query can contain a maximum length \lstinline{n} of the path (\lstinline{node(x).path(y, n)}), i.e. an empty path is returned if a path of size \lstinline{<n} is not found.

These neighborhood and path queries can be used to define features that capture the local graph context in the data. 
For example the size of the neighborhood (\lstinline{neighborsWithin(n).size}) or its diversity (\lstinline{neighborsWithin(n).groupBy(_.tag).size}) may be used to represent an instance, while the length of the path between two instances (\lstinline{path(y).size}) captures the similarity.
It should be noted that such queries can get computationally very complex in large graphs. In order for such queries to be efficient, this function is implemented using breadth-first search over the nodes in the graph. More sophisticated optimization techniques are to be investigated in the future.  


\subsection{Aggregation functions}
We can apply various \emph{aggregation functions} on a collection of properties: 
\begin{lstlisting}
 propertyName.aggregationFun()
\end{lstlisting}
Since the property values have Scala's basic types, a large number of aggregation functions are available to \Saul via built-in Scala functionality. 
Hence, depending on the type of the collection, various functions may be applied.
In case of numeric property types, for example, we can apply aggregations such \lstinline{sum}, \lstinline{product}, and \lstinline{max}. 
A number of aggregation functions are available for any type of instances, such as \lstinline{size} and \lstinline{mkString} for customized concatenation, and further, users can also implement their own aggregation functions by using \lstinline{filter}, \lstinline{map}, \lstinline{reduce}, and \lstinline{groupBy}. In \Saul, such native Scala aggregation functions can be applied on the nodes and properties of the graph.

\section{Querying for learning models}
\label{sec:queryForLearning}
One main goal of basing \Saul\ around a graph based data modeling language is to provide the facility and flexibility for relational learning models to extract complex features. 
Formally, in supervised learning, given a set of examples i.e. pairs of input and output, $E=\{(\x^{(i)},\y^{(i)})\in \mathcal{X}\times \mathcal {Y}: i =1\dots N\}$, learning is defined as a mapping  $h:\mathcal{X}\mapsto\mathcal{Y}$. In general both inputs ($\mathcal{X}$) and outputs ($\mathcal{Y}$) can be arbitrary complex structures. 
In \Saul both $\x$ and $\y$ are sub-graphs of the \datamodel. Each input $\x$ is a set of nodes $\{x_{1},\hdots,x_{K}\}$ and each node has a \emph{type} $p$. Each $x_{k}\in \x$ is described by a set of properties relevant to its type; this set of properties will be converted to a feature vector $\phi_{p}$. For instance, an input type can be a word (\textit{atomic node}) or a pair of words (\textit{composed node}), and each type is described by its own features (e.g. a single word by its part of speech, the pair by the distance of the two words).
The output space $\y$ is represented by a set of \emph{labels}\  $\boldsymbol{l}=\{l_{1},\hdots,l_{P}\}$ each of which is another property of a node in the graph. The labels can have semantic relationships to each other, so that they can represent complex output concepts for any arbitrary task. 
The main components that must be declared for the learning models are learning examples $\x$ and $\y$ and the features. In structured output prediction tasks the background knowledge about the output space should also be declared. Here, we describe how our graph data modeling and various graph queries are used in defining the components of the learning models. 

\subsection{Learning examples}
\label{Learning examples}

Classically, each machine learning example has an input which has a feature vector representation and an output label which is a single variable. A label can be a binary or a multivalued variable in the classification setting, and a real valued variable in the regression setting. 
Classic learning models are the basic building blocks for composing \Saul{'s} complex learning model configurations.  

\subsubsection{Example representation.}
In our integrated feature extraction and learning environment, each learning example is a rooted sub-graph. Each learning model is applied on a specific root node. The input feature types and the label type of the learning examples are a set of graph queries returning direct or contextual properties of the root node. In other words, the root node is the pivot of all the queries that are used for extracting features for a learning example. Consequently, we can define a \textit{learning example} with a node and a set of pivoted queries, one of which is the label query and the rest of which are feature queries. All these queries return a property or a collection of properties.  

The signature for defining a learning model, called \lstinline{modelName}, is as follows: 
\begin{lstlisting}
object modelName extends Learnable(rootNodeName) {/* body */}
\end{lstlisting}
where \lstinline{rootNodeName} is the name of the root node (in the typed graph) of the examples that the learning model takes. Different kinds of properties related to the root node or its connected nodes can be used to define features in the form of queries. These queries define the feature types and the label type of this learning model. As described in Section~\ref{Graph Queries}, queries can be declared and named as properties in the \datamodel and used in the body of the learners or they can be declared directly when defining the learning model.
The following snippet shows how the feature information can be provided in the body of the learning model:
\begin{lstlisting} 
def label = queryLabel
def feature = using(query1,query2,...)
\end{lstlisting}
In this snippet we assume all queries are declared and the learner refers to a list of query names to be used as features.   

\subsubsection{Example selection.}

The examples used by a learner are instances of a single node in the graph and can be retrieved from the graph. It is often the case that the programmer needs to filter the data items and use only a subset of examples for training. A common use case is filtering the negative examples in an unbalanced data set. This can be declared as a part of the learning model. Alternatively, it can be defined elsewhere and set as the example selection filter for the learning model when populating the training data into the data graph. The default filter uses all instances of the learner's type that exist in the instantiated graph. 

\subsubsection{Constraints.}
Constraints may be used in structured learning models to incorporate domain knowledge by explicitly linking some output predictions. Constraints are implemented using queries to specify the relevant properties in the data graph and using a logical expression to impose restrictions on the sets of values they can take. For example, 
\begin{lstlisting}
(sentences(s)$\sim$>sentenceToPhrase)._forall{x=> isPredicate on x is "True" ==> isArgument on x isNot "True"
\end{lstlisting} 
indicates that all phrases in a sentence can be labeled as a predicate or as an argument not both. 

The constraints are defined in terms of the outputs of the classifiers and their primary usage is to impose structural restrictions on the output values of classifiers while making joint predictions. The constraints can be used for structured learning models for joint training too. The details about the underlying computational model for using global constraints in learning and prediction in \Saul is provided in~\cite{KordjamshidiRoWu15}. A constrained classifier can always be used in the body of the properties to generate the property values of the nodes in the \datamodel and called via graph traversal queries. This implies, the answer to such queries is found by performing global inference and finding the best assignments under the user specified constraints.  

\section{Biological networks Application}
\label{sec:bioExample}
\begin{figure}
    \centering
        \includegraphics[trim=0.8cm 1.2cm 0cm 1.2cm, scale=0.42,clip=false]{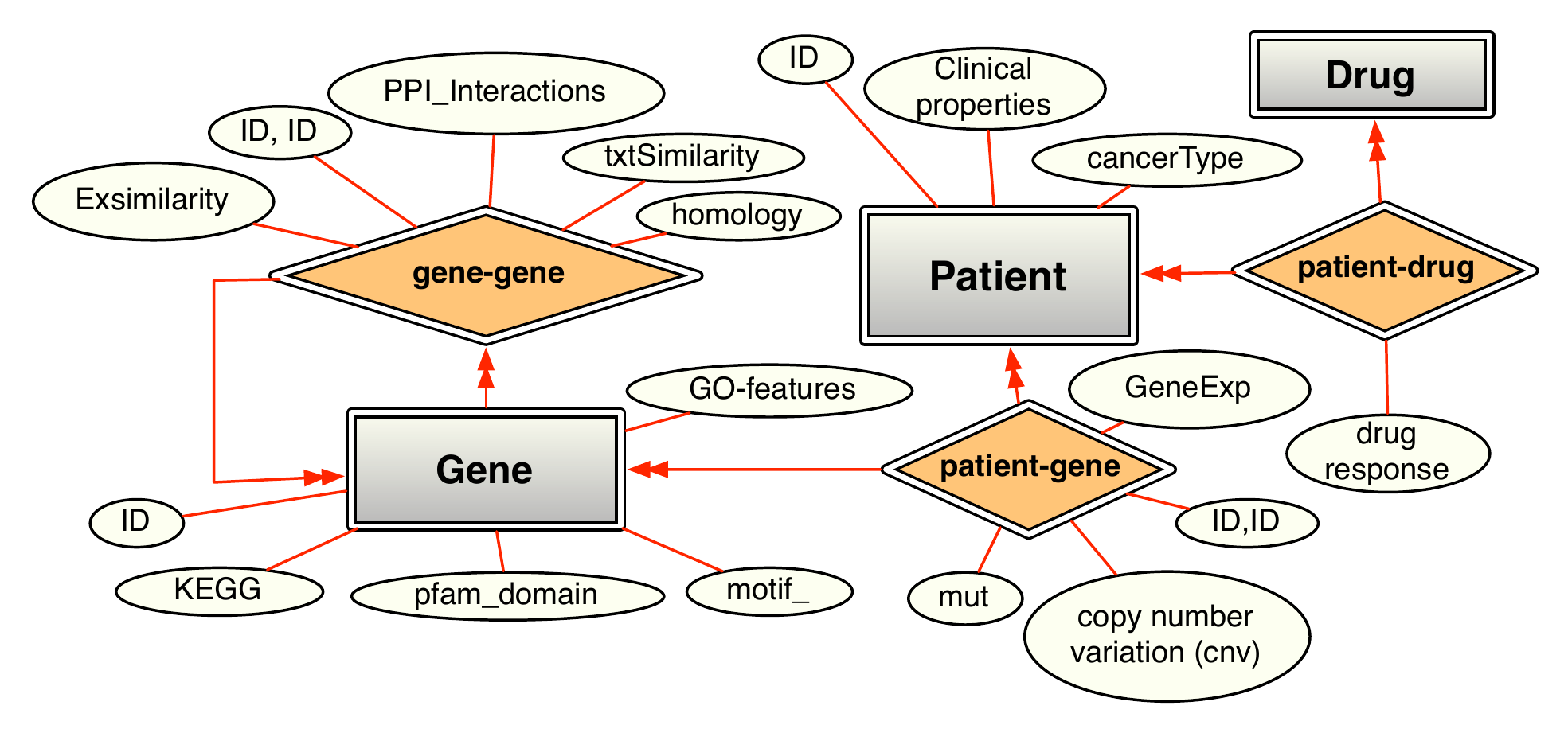}
    \vspace{-0.0cm}
    \caption{This graph represents the type of entities and relationships that are involved in our biological network.}
    \vspace{-0.2cm}
    \label{fig:DataModel_knowEng}
\end{figure}

In this section, we show the value in integrating heterogeneous data coming from various resources in a unified data model for the computational biology domain and the way it facilitates performing various kinds of analysis by querying, learning and inference.  Figure~\ref{fig:DataModel_knowEng} shows the conceptual model of the employed data in terms of entities and relationships. In \Saul all the entities and relationships are declared as nodes in a graph and there is no distinction between them. 
\subsection{Data description}
Our experimental data\footnote{see KnowEng project \url{https://knoweng.org/research/analytics/}} includes various spreadsheets. One spreadsheet is about patients, including an identifier for each patient and some clinical properties of the patients such as age, race and type of cancer. This data is shown in the diagram with the patient box and the type of properties of the patients are connected to this box. There is another spreadsheet that contains the response of each patient to different drugs. This is shown in the patient-drug diamond (relation symbol in relational data models) in the figure. The drug response is a real-valued property of this relation. The genomic information of the patients shown with the patient-gene diamond. This is a spreadsheet that shows the gene expression levels in each patient. The actual data contains the expression values of $\sim20k$ genes, typically for a few thousands of patients.  
The left part of the diagram shows the schema of our biological knowledge graph, which contains the information about genes and their relationships. This is used as background knowledge when performing analysis on patients' data. 
The Gene box shows the individual properties of each gene, such as the pathways that it belongs to, biological processes (from Gene Ontology) it is active in, etc. The Gene-Gene diamond shows relationships between genes, e.g., their sequence similarity, their protein-protein interactions, etc. 

\subsection{Data model declaration}
Given the conceptual model of the data, we can declare the \datamodel at the same conceptual level using Algorithm 1: 
\begin{lstlisting}
 val patients = node[Patient]
 val genes = node[Gene]
 val patientGene = node[PatientGene]
 val patientDrug = node[PatientDrug]
 val geneGene = node[GeneGene]
 val geneGenes = edge(geneGene, genes)
 ...
\end{lstlisting}

\noindent The properties are assigned to nodes, e.g., 
the response of a patient to a drug 
given base-types and sensors, is defined as,

\begin{lstlisting}
val drugResponse = property(patientDrug) { x: PatientDrug => x.response } 
\end{lstlisting}
Similarly the KEGG property is defined as the set of pathways (catalogued in the public KEGG database) that are known to contain that gene,
\begin{lstlisting}
val KEGG=property(gene){x:Gene=>x.pathways} 
\end{lstlisting}

\subsection{Querying the biological network}
Given the properties, genes can be grouped by the pathways. This results in lists of genes per pathway. 
\begin{lstlisting}
val geneGroupedPerPathways = genes.SGroupBy(KEGG,GeneName)
\end{lstlisting}
The \lstinline{geneGroupedPerPathways} is a mapping between pathway names and lists of gene names. Using this mapping we can get a set of genes using a pathway name as a key. 
\begin{lstlisting}
val myPathwayGenes = geneGroupedPerPathWays.get("hsa01040")
\end{lstlisting}
This would return all the genes in our genomic data that belong to pathway ``hsa01040". Biologists might need to retrieve the subsets of genes that belong to a specific pathway as described above and seek all the genes connected to this subset by a specific type of edge. For example, one edge in our experimental knowledge graph is \lstinline{PPIBioGrid} edge (protein-protein interaction edge catalogued in the BioGrid database). To obtain all genes connected by such an edge type to any gene in a specific pathway, the following query can be made: 

\begin{lstlisting}
val pathwayNeighbors = myPathwayGenes.map(gen =>(genes(gen)$\sim$> -geneGenes).filter(rel=> PPIBioGrid(rel).equals(1)))
\end{lstlisting}
This query maps each gene in \lstinline{myPathwayGenes} to all the genes that are connected to it via \lstinline{geneGenes} relation inverse edge, then filters these pairwise gene relations to be limited to a specific class of relations, viz., all relations whose \lstinline{PPIBioGrid} property is equal to one. The goal of this query and the previous one is to obtain a specific subset of genes to be used, for example, in learning or in computing the correlation between the gene expression values and drug responses across all patients. Towards this goal, one may obtain the expression values of a given subset of genes after usage of a specific drug, by defining a property based on joining patient-drug and patient-gene information, as follows,
\begin{lstlisting}
val PWgeneExpression=property(patientDrug){ 
  pd: PatientDrug => 
  patientGene().filter(_.pid ==_.pid).
  filter(myPathWayGenes.contains(GeneName(_))).
  map(_.gExpression)  }
\end{lstlisting}
This definition, given a patient identifier, it retrieves its gene expression measurements and filters them using the subset of genes selected in the previous query above. 

\subsection{Learning and regression} 

A user can define a machine learning model that uses the just-obtained subset of gene expression values for each patient and measures its correlation with the drug response, using a multi-regressor. The multi-regressor is a learner that is defined as described in Section~\ref{Learning examples}:

\begin{lstlisting}
object DrugResponseRegressor extends Learnable(patientDrug){  
def label = drugResponse override   
def feature = using(PWgeneExpression) 
def classifier = new StochasticGradientDescent()  }
\end{lstlisting}

\noindent This learner defines the expression of the above-mentioned subset of genes as the input features and the drug response as the target label. 
The user can upload the data into the knowledge graph as: 
\lstinline{KnowEngDataModel.populateWithKnowengdata()}. 

\noindent This will read the data from files to nodes and edges in the graph. Then the data for each patient will serve as training examples and the regressor can be trained and then tested easily with a few lines of code as follows:   

\begin{lstlisting}
DrugResponseRegressor.learn(
    patientDrugTrainingInstances)
DrugResponseRegressor.testContinuous(
    patientDrugTestingInstances)
\end{lstlisting}

\noindent The \lstinline{testContinuous} method 
reports the regressor's evaluation using various metrics including sum-square of residuals, Pearson correlations, etc. 

\subsection{Meta analysis based on learning results}
\Saul facilitates meta analysis of the results of various learning models. In the previous problem, the user can easily define a set of learners by parameterization of the input properties. For example, a property (called \lstinline{PWgeneExpression}) can receive the name of a pathway as an input parameter and return the expression of the genes of that specific pathway for a specific patient. 

\begin{lstlisting}
val pathWayGExpression= (pathway: String) => 
  property(patientDrug, ordered = true) {
  pd: PatientDrug =>
    val myPathwayGenes = genesGroupedPerPathway.get(pathway) 
    patientGene().filter(_.pid == _.pid).
    filter(myPathwayGenes.contains(GeneName)).
    map(_.gExpression) }
\end{lstlisting}
Assume that we have 100 pathways in our data graph and the biologist wants to define a separate regressor each of which uses the genes of each pathway as input features. This can be done by defining a class of learners, parameterized with the \lstinline{pathway} property: 

\begin{lstlisting}
class DrugResponseRegressor(pathway: String) extends Learnable(patientDrug){  
    def label = drugResponse 
    def feature = using(PWgeneExpression(pathway))  }
\end{lstlisting}

\noindent Given this class of models, we can create as many regressors as the number of distinct pathways in our data, 

\begin{lstlisting}
val myLearners=(genes() prop gene_KEGG).
    flatten.distinct.map(new DrugResponseRegressor(_))
\end{lstlisting}

\noindent \lstinline{myLearners} is a collection of regressors, each of which is created based on a set of gene expression values corresponding to a specific pathway. Now we can apply train, test, etc on all learners on this collection at once: 
\begin{lstlisting}
myLearners.map(_.train())
val testRslts = myLearners.map(_.test()) 
val sortedRgrs = testRslts.SortWithAccuracy
val bestRegressor = testRslts.maxAccuracy
\end{lstlisting}
The first line in the above snippet trains all the regressors. The second line shows how to test the regressors according to a specific evaluation metric, for example accuracy. 
For a different task on the same data, the user can seamlessly define new nodes, properties and classifiers and reuse the existing data model. 
%

\section{Related Work and Discussion}
\label{RelatedWork}
Our proposed model is related to many existing systems from various perspectives. We highlight the differences and similarities and the new advantages of our \datamodel in the context of \Saul~\cite{KordjamshidiRoWu15}. 
\\\noindent{\bf Comparison to machine learning tools.}
Most of the commonly used ML tools such as WEKA~\cite{Witten99weka:practical} or Mallet~\cite{McCallumMALLET} 
provide easy access to learning algorithms. However, a common characteristic of these tools is that a flat data structure in a specific file format should be provided. 
This is a major disadvantage for relational learning when a) the data domain is structured and features should be extracted from parts of the structure  b) there are several learning models involved that interact with each other and use multiple feature generation tools (potentially from different sources) c) the user needs to do experimentation and feature engineering, which is often the case when designing machine learning models. 
One goal of \Saul's data modeling language is to address these issues on top of machine learning models. 

\noindent{\bf Comparison to feature extraction languages.} Feature extraction is very challenging in relational data domains 
such as computational biology, or when the data is raw with a complex and implicit structure such as natural language or computer vision. There are some tools available in the latter two domains which facilitate feature extraction by providing generic data structures appropriate for those domains and set of tools applicable on those data structures (i.e. readers and sensors). A recent example in NLP domain is Fextor~\cite{fextor}. This tool provides an internal representation for textual data and provides a library to make queries, relying on its fixed internal representation. 
Prior to Fextor, Fex~\cite{CumbyRo03,CumbyRo00} views feature extraction from a first order knowledge representation perspective, and it is closer to our view here. 
However, their formalization is based on Description Logic~\cite{baader2003description} where each feature extraction query is answered by logical reasoning. While having similar perspective, 
we solve queries using graph traversal over the propositionalized graph instead of logical reasoning. 
We are able to run the same type of queries given our graph-based formulation~\cite{Shastri91}.
Unlike Fex, our data model declaration provides the flexibility of working with arbitrary types of objects--e.g. we are not limited to having sentence level features, or tokens level features, etc. 
 \Saul's data modeling language enables the user to declare the graph and plugging in any arbitrary data structures and arbitrary external sensors. 


\noindent{\bf Information extraction tools.} With the ability to work on unstructured data, 
our system has many common features with the information extraction tools. To facilitate working on unstructured data, there has been efforts in designing unified data structures for processing textual data and preparing tools (i.e. sensors) that can operate on those data structures~\cite{sammons2016edison}. A well-known example of such universal data structure is UIMA~\cite{ferrucci2004uima} that can be augmented with NLP  tools. Similarly, there are some well-known software that focus on providing NLP sensors, such as NLTK~\cite{Loper:2002:NNL:1118108.1118117}, GATE~\cite{Cunningham2011a}. 
These frameworks, focus on providing a specific internal representation and  do not allow for a declaration of a model based on arbitrary structures and using arbitrary external sensors easily in one data model. Though some of these information extraction systems are equipped with very well designed and efficient query languages such as SystemT~\cite{KLRRVZ09}, 
we argue for a generic framework for information extraction in the context of heterogeneous information networks while addressing learning and inference in a same framework.  


\noindent{\bf Relational and graph based query languages.} The concept of graph queries and using first order logical languages for querying from structured data is well-established for relational and graph database technologies~\footnote{http://tinkerpop.incubator.apache.org/}. 
Along the same line, our data modeling language facilitates querying form structured data. Graph traversal approaches are known to be more efficient for performing join operations and there are scalable implementations 
available for working on graph structures
~\cite{186216}. Our goal is integrating such capabilities with learning based programming.  



\noindent{\bf In the context of \Saul.} 
Our proposed feature language currently implemented as a part of 
\Saul~\cite{KordjamshidiRoWu15}, that does structured-learning based on Constraint Conditional Models~\cite{chang2012structured} but can be integrated with any JVM-based languages which are designed for advanced machine learning models such as probabilistic graphical models e.g. WOLFE~\cite{riedel2014wolfe} and FACTORIE~\cite{McCallumScSi09}. Using such a data modeling language, structured output prediction models can exploit the expressive power of relational feature representation to easily handle the issues of representation of the structured inputs and outputs, and define relational features over them.
In contrast to our proposed \datamodel, the aspect of data modeling and feature extraction is less elaborated in other relational learning frameworks. For example, in Alchemy for programming Markov Logic Networks (MLNs)~\cite{RichardsonDo06}, raw data should be processed offline and stored in a DB file in predicate-argument form. It is flexible only in the way that first-order logical expressions can be written to operate on the predefined predicates retrieved from the DB. The relational learning language kLog~\cite{frasconi2014klog} has the same input structure and format. However, kLog is more flexible from the feature extraction point of view, because it uses Prolog and provides the possibility of logical reasoning for feature extraction. These systems are not designed 
to be integrated with various sources of information and building an information network is not their concern. 
The feature extraction is treated as an external prepossessing component in such frameworks. 


\section{Conclusion}  
\label{sec:conclusion}
In this work we propose an initial prototype for a new integrated graph-based data-modeling and feature extraction language as a part of declarative learning language, \Saul, to facilitate building end-to-end learning based systems for real world applications. We address the issue of building a heterogeneous information network based on both structured and unstructured data from various resources. Our language combines the power of relational, graph-based feature extraction, with the flexibility to exploit previously established resources such as data readers, annotators, sensors, and data structures in a given domain. We describe examples in NLP and computational biology domains. 
In summary our prototype, similar to the existing data modeling languages, provides the capability of declarative querying from data using relational operations; in contrast to existing systems, a) provides the possibility of seamless integration of unstructured and heterogeneous data from various domains into a unified data model; b) connects the relational graph directly to the analysis units which are relational machine learning models; c) facilitates on-the-fly integration of the output of the analysis units into an evolving graph.

\section*{Acknowledgements}
The authors would like to thank all the students who have helped in the implementation of this project, as well as the anonymous reviewers for helpful comments.  This research is supported by NIH grant U54-GM114838 
awarded by NIGMS through funds provided by the trans-NIH Big Data to Knowledge (BD2K) initiative (www.bd2k.nih.gov). This material is based on research sponsored by DARPA under agreement number FA8750-13-2-0008. The U.S. Government is authorized to reproduce and distribute reprints for Governmental purposes notwithstanding any copyright notation thereon. The views and conclusions contained herein are those of the authors and should not be interpreted as necessarily representing the official policies or endorsements, either expressed or implied, of DARPA or the U.S. Government nor NIH.

\bibliography{new,ccg,cited}
\end{document}